\documentclass[conference]{IEEEtran}
\IEEEoverridecommandlockouts
\usepackage{cite}
\usepackage{amsmath,amssymb,amsfonts}
\usepackage{algorithm}
\usepackage{algorithmic}
\usepackage{float}
\usepackage{graphicx}
\usepackage{textcomp}
\usepackage[table]{xcolor}
\usepackage{booktabs}
\usepackage{multirow}
\usepackage{makecell}
\usepackage{xurl}
\usepackage[hidelinks]{hyperref}
\usepackage{flushend}
\def\BibTeX{{\rm B\kern-.05em{\sc i\kern-.025em b}\kern-.08em
    T\kern-.1667em\lower.7ex\hbox{E}\kern-.125emX}}
\newcommand{\LEGO}{\textsc{Lego}}

\newsavebox{\expOneTableBox}
\newsavebox{\expTwoTableBox}
\definecolor{PriorWorkCol}{HTML}{F6E6C7}

\begin{document}

\title{\raisebox{-0.15\height}{\includegraphics[height=0.95em]{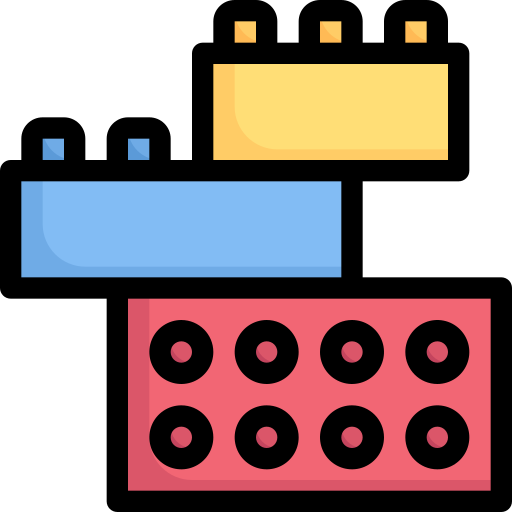}}\hspace{0.35em}\LEGO{}: An LLM Skill-Based Front-End Design Generation Platform
}

\author{%
\IEEEauthorblockN{Jincheng Lou\textsuperscript{1}, Ruohan Xu\textsuperscript{2}, Jiecheng Ma\textsuperscript{3}, Runzhe Tao\textsuperscript{1}, Xinyu Qu\textsuperscript{1}, and Yibo Lin\textsuperscript{1,4,5,*}}
\IEEEauthorblockA{\textsuperscript{1}School of IC, Peking University, \textsuperscript{2}School of EECS, Peking University,\\
\textsuperscript{3}School of Microelectronics, Xidian University, \textsuperscript{4}Institute of EDA, Peking University,\\
\textsuperscript{5}Beijing Advanced Innovation Center for IC}
\IEEEauthorblockA{Email: jinchenglou@stu.pku.edu.cn \textsuperscript{*}Corresponding author: yibolin@pku.edu.cn}
}

\maketitle

\begin{abstract}
Existing LLM based EDA agents are often isolated task-specific systems. This leads to repeated engineering effort and limited reuse of successful design and debugging strategies.
We present \LEGO{}, a unified skill-based platform for front-end design generation. It decomposes the digital front-end flow into six independent steps and represents every agent capability as a standardized composable circuit skill within a plug-and-play architecture.
To build this skill library, we survey more than 100 papers, select 11 representative open-source projects, and extract 42 executable circuit skills within a six-step finite state machine formulation.
Circuit Skill Builder automates skill extraction with linear scalability. Agent Skill RAG achieves submillisecond retrieval without relying on embedding models.
Empirical evaluation on a hard subset of 41 VerilogEval v2 problems that gpt-5.2-codex fails to solve under extra-high (xhigh) reasoning effort shows that individual circuit skills constructed within \LEGO{} raise Pass@1 from 0.000 to 0.805. This is an 80.5\% gain over the baseline.
Cross project skill compositions also reach 0.805 Pass@1. They outperform \texttt{hierarchy-verilog} by 14.6\% and \texttt{VerilogCoder} by 2.5\%. They also match \texttt{MAGE}. These results show that modular skill composition supports both effective and flexible RTL design automation.
The \LEGO{} platform and all circuit skills are publicly available at GitHub (\url{https://github.com/loujc/LEGO-An-LLM-Skill-Based-Front-End-Design-Generation-Platform}).
\end{abstract}

\begin{IEEEkeywords}
LLM Agents, Agent Skills, RTL Generation, Front-End Design Flow, VerilogEval
\end{IEEEkeywords}

\section{Introduction}

Automating the digital front-end design flow is a central challenge in AI assisted chip design.
The flow covers specification refinement, RTL coding, testbench generation, simulation, and iterative debugging.
Generated RTL code must satisfy strict syntax rules, timing constraints, and signal-level functional correctness.
This breadth and the stringent requirements at each step make end-to-end automation difficult and expensive.
\begin{figure}[ht]
\centering
\includegraphics[width=\columnwidth]{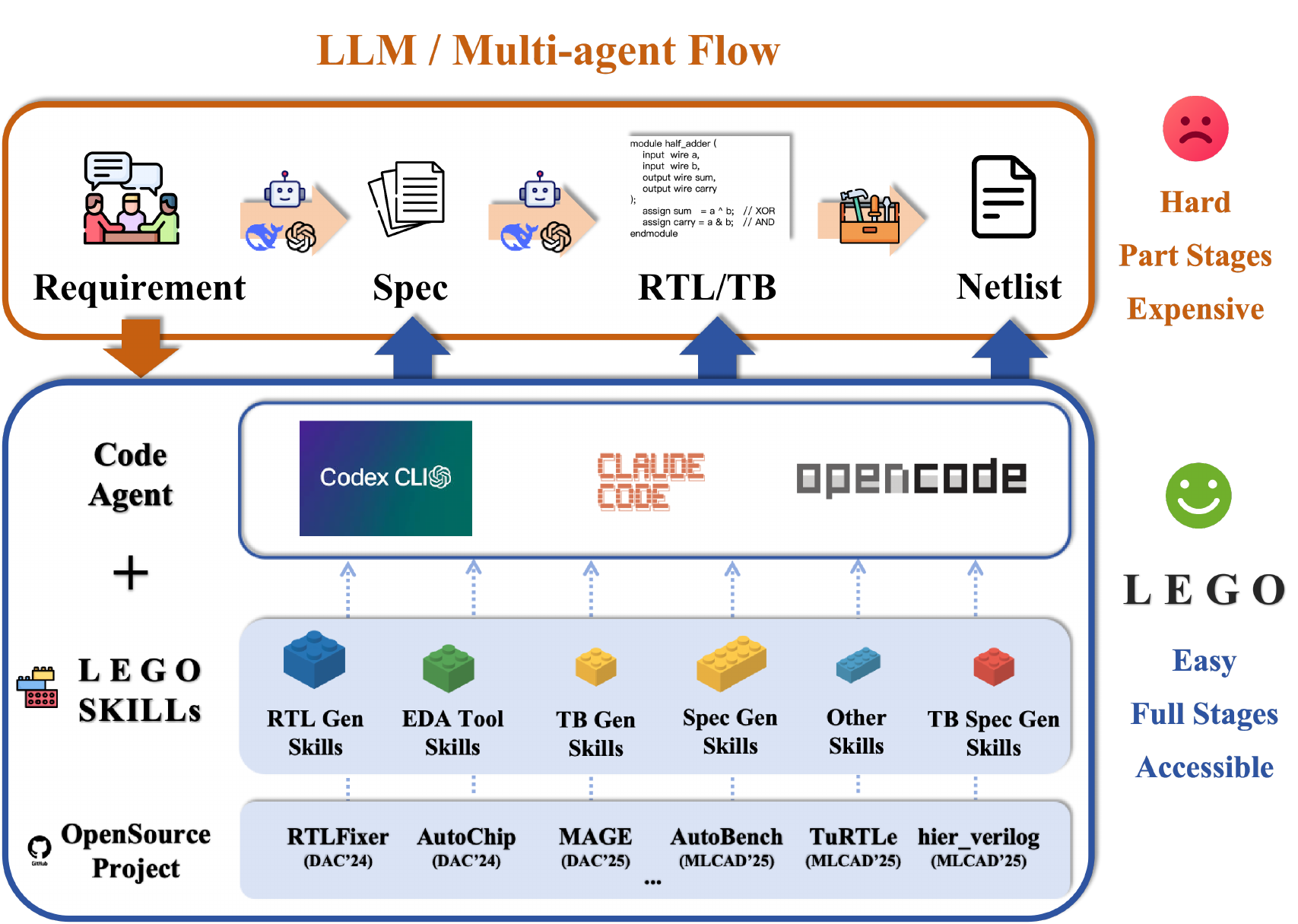}
\caption{\LEGO{} System Overview}
\label{fig:intro-overview}
\end{figure}
Large language models and code agents have shown strong potential in this domain. Prior work has made progress in benchmark construction~\cite{liu_verilogeval_2023,pinckney2024revisitingverilogevalnewerllms,10473904,11310982,11126373}, agent-based design workflows~\cite{liu2024chipnemodomainadaptedllmschip,nadimi2025verimindagenticllmautomated,Ho_Ren_Khailany_2025,11105957,11133191}, and domain-specific LLM training~\cite{11145160,10720939,kumar2024hdlgpthighqualityhdlneed,10.1145/3643681,11126250,10817982}. Most existing systems are isolated solutions. They bind algorithmic ideas to project-specific toolchains. This limits cross-project reuse and forces repeated implementation of common infrastructure.

To address this fragmentation, we propose \LEGO{}, a unified skill-based front-end design generation platform.
Fine-tuning approaches and multi-agent systems often reimplement infrastructure for partial flows and incur high LLM API cost. \LEGO{} instead decomposes the workflow into six-steps and supports plug-and-play composition across steps.
As shown in Fig.~\ref{fig:intro-overview}, \LEGO{} runs on top of code agent tools. It can therefore use vendor provided coding plans. These plans are more affordable and accessible than raw LLM API calls.
Circuit Skill Builder converts open-source projects into \LEGO{} compliant circuit skills within the open code agent ecosystem. Agent Skill RAG accumulates design and debugging experience for efficient reuse.

Our contributions are listed below.
\begin{itemize}
\item \textbf{\LEGO{} is the first open skill-based platform for digital front-end design.} It decomposes the workflow into six independent steps and supports plug-and-play composition. This enables cross-project reuse without rebuilding infrastructure. The platform and extracted circuit skills are publicly released in a unified GitHub repository.
\item \textbf{Executable skill library.} We curate a library of 42 executable circuit skills extracted from 11 representative open-source projects and organized into 24 functional groups. Circuit Skill Builder automates skill extraction. Agent Skill RAG supports lightweight experience accumulation and achieves submillisecond retrieval.
\item \textbf{Strong empirical gains.} We validate the framework on a hard subset of 41 VerilogEval v2 problems from tasks that gpt-5.2-codex fails to solve. The results show an 80.5\% Pass@1 improvement over the baseline. Composed skills outperform hierarchy-verilog by 14.6\% and VerilogCoder by 2.5\%. They also match the RTL generation SOTA method \texttt{MAGE}.
\end{itemize}

\section{Preliminary}

\subsection{LLM and Multi-Agent Paradigms in RTL Generation}

Compared with software languages, Verilog and other HDLs encode hardware hierarchy, timing behavior, and signal dependencies in a tightly coupled form.
RTL generation therefore requires coordinated reasoning across specification understanding, code synthesis, simulation feedback, and iterative correction, making the task difficult for a single static prompting strategy.

Current approaches to LLM assisted RTL generation follow two main directions.
The first direction trains domain-specific LLMs on hardware description data~\cite{11145160,10720939,kumar2024hdlgpthighqualityhdlneed,10.1145/3643681,11126250,10817982}.
However, the scale of publicly available RTL datasets is orders of magnitude smaller than that of software code. The structural gap between HDL and general-purpose programming languages further limits transfer.
These factors make domain-specific models expensive to train and their performance still lags behind expectations.
The second direction leverages multi-agent systems~\cite{liu2024chipnemodomainadaptedllmschip,nadimi2025verimindagenticllmautomated,Ho_Ren_Khailany_2025,11105957,11133191}. These systems benefit directly from the rapid improvement of general-purpose LLMs and have become the mainstream approach.
Multi-agent systems decompose the workflow into specialized roles with step-specific tools and prompts.
This decomposition reduces context interference across heterogeneous subtasks such as specification analysis, RTL coding, testbench generation, and debug.
Our framework follows this collaborative principle and further standardizes each executable capability as a reusable skill unit.

\subsection{Code Agents and the Abstraction of Skills}

A code agent is an autonomous programming system that can invoke external tools and react to execution feedback.
Representative code agents include Codex CLI~\cite{openai_codex_cli}, Claude Code~\cite{anthropic_claude_code}, and OpenCode~\cite{opencode}.
Beyond token generation, these agents execute shell commands, interpret compiler and simulator outputs, and perform iterative error correction.
The digital front-end design flow is predominantly composed of command-line invocations, structured text processing, and code manipulation, making code agents a natural fit for this domain.

These code agent platforms provide an official abstraction called Skill as an atomic capability unit for environment interaction and structured reasoning.
A skill encapsulates a single well-defined capability such as generating RTL code, running simulation, or extracting error messages. It acts as a self-contained module with clearly defined inputs, outputs, and execution logic. This enables composition, reuse, and independent updates.
Consistent with the seven field skill specification in Section~\ref{subsec:method-skill-lib}, the formal representation of a skill is
\begin{equation}
S = (N, F, C, P, IO, \Sigma, G)
\label{eq:skill}
\end{equation}
where $N$ is the skill name, $F$ is the simple function description, $C$ is the constraints section, $P$ is the entrypoint command, $IO$ is the input and output specification, $\Sigma$ is the schema definition, and $G$ is the done criteria. This abstraction provides a consistent interface for composition, execution, and experience accumulation.

\section{Methodology}

\begin{algorithm}[t]
\caption{Circuit Skill Builder}
\label{alg:skill-builder}
\begin{algorithmic}[1]
\STATE \textbf{Input} Project $P$, paper $R$, and taxonomy $\mathcal{T} = \{w_1, \ldots, w_6\}$
\STATE \textbf{Output} Skill set $\mathcal{S}_{\text{new}}$ and skill groups $\{\mathcal{G}_j\}$
\STATE
\STATE \textbf{Step 1. Summarize} // Extract capabilities
\STATE $\mathcal{M} \gets$ Summarize innovations and functions from $P$ and $R$ using LLM
\STATE
\STATE \textbf{Step 2. Extract} // Map to workflow steps
\STATE $\mathcal{S}_{\text{cand}} \gets \emptyset$
\FOR{each capability $m \in \mathcal{M}$}
    \STATE Identify target step $w_i \in \mathcal{T}$ for $m$
    \STATE Generate $(N, F, C, P, IO) \gets$ Extract fields from $m$ and $P$ codebase
    \STATE $\mathcal{S}_{\text{cand}} \gets \mathcal{S}_{\text{cand}} \cup \{(N, F, C, P, IO)\}$
\ENDFOR
\STATE
\STATE \textbf{Step 3. Post process} // Standardize and organize
\FOR{each $s \in \mathcal{S}_{\text{cand}}$}
    \STATE Complete $s$ with schema $\Sigma$ and done criteria $G$
\ENDFOR
\STATE $\mathcal{S}_{\text{new}} \gets \text{Deduplicate}(\mathcal{S}_{\text{cand}})$
\STATE Classify $\mathcal{S}_{\text{new}}$ into $\mathcal{S}_1, \ldots, \mathcal{S}_6$ by workflow step
\STATE Compose groups $\{\mathcal{G}_j\}$ based on functional complementarity
\STATE
\RETURN $\mathcal{S}_{\text{new}}$, $\{\mathcal{G}_j\}$
\end{algorithmic}
\end{algorithm}

\begin{table*}[t]
\centering
\caption{Circuit Skill List and Group Notations in \LEGO{}}
\label{tab:skills-detail}\label{tab:skills-summary}
{\scriptsize
\setlength{\tabcolsep}{5pt}
\renewcommand{\arraystretch}{1.25}
\centering
\begin{tabular}{@{} l l c >{\raggedright\arraybackslash}p{0.63\textwidth} @{}}
\toprule
\rowcolor{black!8}
\textbf{Step Skills} & \textbf{Group} & \textbf{\#} & \textbf{Included Circuit Skills} \\
\midrule
RTL Spec & S1 & 8 & hier\_submodule\_list\_prompt, hier\_outline\_prompt, hier\_subhier\_json\_prompt, hier\_function\_check\_prompt, hier\_header\_check\_prompt, hier\_refine\_question\_prompt, hier\_refine\_integrate\_prompt, prompt\_enricher \\
 & S2 & 1 & spec2rtl-understanding-pipeline \\
\midrule
TB Spec & TS1 & 1 & autobench\_circuit\_type\_classify \\
 & TS2 & 1 & autobench\_tb\_scenarios\_prompt \\
 & TS3 & 1 & autobench\_tb\_rules\_extract \\
 & TS4 & 1 & autobench\_tb\_spec\_prompt \\
 & TS5 & 1 & iverilog\_waveform\_parser \\
 & TS6 & 1 & mage\_sim\_judge\_tb\_fix \\
 & TS7 & 1 & verilogcoder\_case\_loader \\
\midrule
RTL Gen & G1 & 1 & verilogcoder\_rtl\_subtask\_prompt \\
 & G2 & 1 & mage\_rtl\_generate \\
 & G3 & 1 & autobench\_rtl\_prompt \\
 & G4 & 2 & hier\_verilog\_gen\_prompt, spec\_ir\_codex\_rtl\_gen \\
\midrule
TB Gen & TG1 & 1 & autobench\_directgen\_prompt \\
 & TG2 & 1 & autobench\_pychecker\_tb\_prompt \\
 & TG3 & 1 & verilogcoder\_tb\_merge \\
\midrule
EDA Tool & E1 & 3 & iverilog\_compile, iverilog\_syntax\_fixer, spec2rtl-closed-loop \\
 & E2 & 5 & iverilog\_compile, iverilog\_error\_localizer, iverilog\_error\_rag, iverilog\_code\_fixer, iverilog\_compile\_fix\_chain \\
 & E3 & 3 & spec2rtl-closed-loop waveform\_feedback, iverilog\_simulator, verilogcoder\_waveform\_trace \\
\midrule
Others & O1 & 2 & iverilog\_error\_localizer, iverilog\_error\_rag \\
 & O2 & 1 & autobench\_runinfo\_analyze \\
 & O3 & 2 & mage\_token\_accounting, mage\_benchmark\_reader \\
 & O4 & 1 & rtl\_fault\_injector \\
 & O5 & 1 & hier\_tree\_ops \\
\midrule
\rowcolor{black!5}
\multicolumn{4}{@{}l}{\textbf{Total:} 6 stages, 24 skill groups, 42 circuit skills} \\
\bottomrule
\end{tabular}
}

\end{table*}

\subsection{Workflow Decomposition and Skill Curation}

As shown in Fig.~\ref{fig:intro-overview}, \LEGO{} decomposes the digital front-end design flow into six sequential steps. These steps are RTL Spec Generation, Testbench Spec Generation, RTL Generation, Testbench Generation, EDA Tool, and Others. They cover specification refinement, code synthesis, compilation, simulation with tools such as Icarus Verilog~\cite{icarus_verilog} and Verilator~\cite{verilator}, and auxiliary utilities.

We model this workflow as a finite state machine. Let $\mathcal{W} = \{w_1, \ldots, w_6\}$ denote the workflow steps, and let $\mathcal{S}_i$ denote the skill set at step $w_i$. The state at iteration $t$ is
\begin{equation}
\Psi^{(t)} = \left( w_i, s_j^{(i)}, \mathcal{A}^{(t)}, \mathcal{D}^{(t)} \right)
\label{eq:workflow-state}
\end{equation}
where $w_i \in \mathcal{W}$ is the current step, $s_j^{(i)} \in \mathcal{S}_i$ is the active skill, $\mathcal{A}^{(t)}$ are accumulated artifacts, and $\mathcal{D}^{(t)}$ tracks design decisions. The transition function $\delta$ maps $\Psi^{(t)}$ to $\Psi^{(t+1)}$ and governs step progression based on EDA tool feedback.

To construct the skill sets $\mathcal{S}_i$, we survey over 100 papers on LLM assisted front-end design and select 11 representative open-source projects~\cite{11106091,11133191,11105957,Ho_Ren_Khailany_2025,10.1145/3670474.3685956,11105878,11189228,thakur2024autochipautomatinghdlgeneration,huang2024llmpoweredverilogrtlassistant,11294897,zhang2026understandingmitigatingerrorsllmgenerated}.
From these projects, 42 standardized circuit skills are extracted and organized into $\mathcal{S}_1$ through $\mathcal{S}_6$ across all six-steps as shown in Table~\ref{tab:skills-detail}.
This systematic formulation, combined with skill construction from proven sources, enables up to 70.7\% Pass@1 improvement over the baseline in Experiments 1 and 2.

\subsection{Skill Specification and Construction}\label{subsec:method-skill-lib}

To ensure cross-project compatibility, each circuit skill follows the seven field specification in Eq.~\eqref{eq:skill}. The fields are name $N$, function description $F$, constraints $C$, entrypoint $P$, input and output specification $IO$, schema $\Sigma$, and done criteria $G$.
The naming pattern \texttt{<project\_abbr>\_<function\_abbr>} guarantees uniqueness, while the function description enables the agent to select the appropriate skill at runtime.
The entrypoint $P$ specifies the command-line invocation, $IO$ and $\Sigma$ define data structures and formats, and $G$ determines completion conditions.

Circuit Skill Builder in Algorithm~\ref{alg:skill-builder} automates skill construction from open-source projects.
Applying this pipeline to 11 projects yields 42 circuit skills organized into 24 groups in Table~\ref{tab:skills-detail}.
Skills are grouped by functional complementarity within each workflow step, ensuring similar input/output interfaces for interchangeable composition.
For instance, RTL Gen groups G1 to G4 represent distinct methodologies such as subtask decomposition, direct generation, specification-driven generation, and hierarchical generation. They come from different source projects.
Beyond these circuit skills, \LEGO{} provides Agent Skill RAG to accumulate debugging and design experience for each workflow step.
Unlike traditional vector-based RAG, Agent Skill RAG employs a lightweight two-stage retrieval mechanism.
First, the agent loads only the concise descriptions of all available RAG entries in each skill, maintaining minimal context overhead.
When a relevant entry is identified through semantic matching, the agent then loads the complete knowledge unit with detailed fields such as symptom pattern, root cause, fix strategy, and applicable conditions.
This lazy loading approach eliminates the need for embedding models or reranking while enabling submillisecond initial retrieval and real-time knowledge updates via simple text appends.
Retrieved fixes are injected as explicit prompts to guide subsequent iterations.

\subsection{Hierarchical Skill Architecture}

\begin{figure*}[t]
\centering
\includegraphics[width=0.96\textwidth]{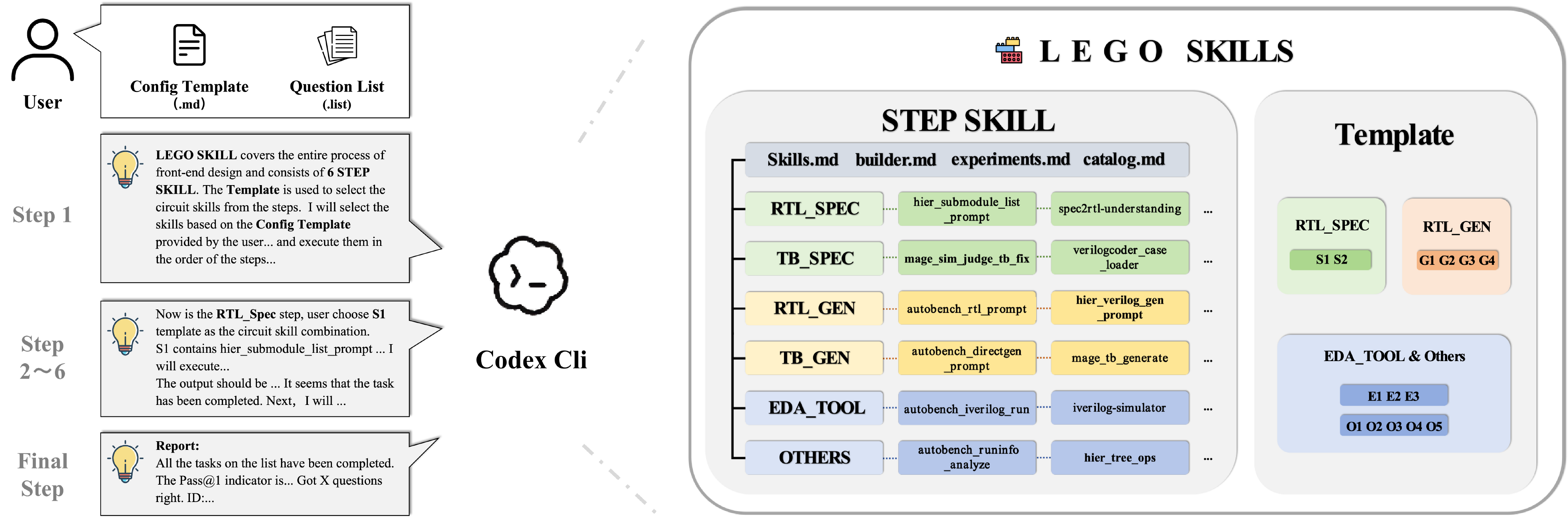}
\caption{Step level view of the \LEGO{} methodology.}
\label{fig:method-step}
\end{figure*}

As shown in Fig.~\ref{fig:method-step}, the skill system of \LEGO{} comprises two components. They are Template and Step Skill.
A Template encapsulates a preset combination of circuit skills, enabling users to initiate design tasks directly without manual configuration.
Within a Template, users can also specify which circuit skills to activate for each step, thereby adapting the system to specific requirements.

The Step Skill component forms a three-layer hierarchy together with the top level \LEGO{} Skill.
The top layer defines the overall workflow decomposition, execution order, and iteration logic.
The middle layer consists of six-step skills, each listing available circuit skills and specifying active configurations dynamically adjusted via the config template.
The bottom layer contains the circuit skills and Agent Skill RAG entries with their complete definitions, forming the atomic execution units of \LEGO{}.

The entire system is lightweight and highly extensible because it consists primarily of markdown files and requires no complex environment setup.
As illustrated in the right panel of Fig.~\ref{fig:method-step}, the execution workflow is straightforward. The user provides a config template and question list. The code agent invokes the \LEGO{} Skill and loads the configuration to select active circuit skills for each step. It then proceeds through the six-steps in order while preserving intermediate results. Finally it generates a summary report upon completion.

\section{Experiments}

\begin{table*}[t]
\centering
\caption{Aggregate metrics of Experiment 1 on 41 filtered VerilogEval v2 problems.}
\label{tab:exp1-summary}
\resizebox{\textwidth}{!}{%
\begin{minipage}{\textwidth}
\centering
\sbox{\expOneTableBox}{{\small
\setlength{\tabcolsep}{6pt}
\renewcommand{\arraystretch}{1.18}
\begin{tabular}{lcccccc}
\hline
\textbf{Metric} & \textbf{gpt-5.2-codex (xhigh)} & \textbf{G1} & \textbf{G2} & \textbf{S1G1} & \textbf{G1E1E2 (w. loop)} & \textbf{G1E3 (w. loop)} \\
\hline
\textbf{Solved} & 0 / 41 & 23 / 41 & 25 / 41 & 22 / 41 & 29 / 41 & \textbf{33 / 41} \\
\textbf{Pass@1} & 0.000 & 0.561 & 0.610 & 0.537 & 0.707 & \textbf{0.805} \\
\textbf{Gain} & +0.0\% & +56.1\% & +61.0\% & +53.7\% & +70.7\% & \textbf{+80.5\%} \\
\hline
\end{tabular}
}
}%
\usebox{\expOneTableBox}
\par\vspace{2pt}
\parbox[t]{\wd\expOneTableBox}{\footnotesize\raggedright\emph{Note.} Gain is computed relative to gpt-5.2-codex under xhigh. Notation such as G1, S1G1, and E1 follows Table~\ref{tab:skills-detail}.}
\end{minipage}%
}
\end{table*}

\begin{table*}[t]
\centering
\caption{Aggregate metrics of Experiment 2 on 41 filtered VerilogEval v2 problems.}
\label{tab:exp2-summary}
\resizebox{0.95\textwidth}{!}{%
\begin{minipage}{\textwidth}
\centering
\sbox{\expTwoTableBox}{{\small
\setlength{\tabcolsep}{4.0pt}
\renewcommand{\arraystretch}{1.18}
\begin{tabular}{lcccccccc}
\hline
\multirow{2}{*}{\textbf{Metric}} & \multirow{2}{*}{\textbf{S1G4}} & \textbf{S1G4E1E2} & \textbf{S1G4E3} & \textbf{S2E1E2E3} & \cellcolor{PriorWorkCol} & \cellcolor{PriorWorkCol} & \cellcolor{PriorWorkCol} & \cellcolor{PriorWorkCol} \\
 &  & \textbf{(w. loop)} & \textbf{(w. loop)} & \textbf{(w. loop)} & \cellcolor{PriorWorkCol}\multirow{-2}{*}{\textbf{hierarchy-verilog~\cite{11105878}}} & \cellcolor{PriorWorkCol}\multirow{-2}{*}{\textbf{verilogcoder~\cite{Ho_Ren_Khailany_2025}}} & \cellcolor{PriorWorkCol}\multirow{-2}{*}{\textbf{MAGE~\cite{11133191}}} & \cellcolor{PriorWorkCol}\multirow{-2}{*}{\textbf{RTLFixer~\cite{11294897}}} \\
\hline
\textbf{Solved} & 21 / 41 & 30 / 41 & \textbf{33 / 41} & \textbf{33 / 41} & \cellcolor{PriorWorkCol}27 / 41 & \cellcolor{PriorWorkCol}32 / 41 & \cellcolor{PriorWorkCol}\textbf{33 / 41} & \cellcolor{PriorWorkCol}17 / 41 \\
\textbf{Pass@1} & 0.512 & 0.732 & \textbf{0.805} & \textbf{0.805} & \cellcolor{PriorWorkCol}0.659 & \cellcolor{PriorWorkCol}0.780 & \cellcolor{PriorWorkCol}\textbf{0.805} & \cellcolor{PriorWorkCol}0.415 \\
\textbf{Gain} & +51.2\% & +73.2\% & \textbf{+80.5\%} & \textbf{+80.5\%} & \cellcolor{PriorWorkCol}+65.9\% & \cellcolor{PriorWorkCol}+78.0\% & \cellcolor{PriorWorkCol}\textbf{+80.5\%} & \cellcolor{PriorWorkCol}+41.5\% \\
\hline
\end{tabular}
}
}%
\usebox{\expTwoTableBox}
\par\vspace{2pt}
\parbox[t]{\wd\expTwoTableBox}{\footnotesize\raggedright\emph{Note.} Gain is computed relative to gpt-5.2-codex under xhigh. Notation follows Table~\ref{tab:skills-detail}.}
\end{minipage}%
}
\end{table*}

This section presents three complementary analyses. Subsection~\ref{subsec:exp-setup} describes the experimental setup, including the model, evaluation metric, and benchmark construction. Subsection~\ref{subsec:exp-step-skill} evaluates the effectiveness of step decomposition and individual circuit skills. Subsection~\ref{subsec:exp-flexibility} studies composed skill pipelines and compares different combinations to validate the flexibility and effectiveness of \LEGO{}.

\subsection{Experimental Setup}\label{subsec:exp-setup}
To ensure that the behavior of each constructed circuit skill remains as close as possible to its source project, we use the official repositories reported in the corresponding papers for all open-source projects except VerilogAssistant. For VerilogAssistant we use an open-source reproduction repository on GitHub~\cite{verilogassistant_github}.
To evaluate the generality and effectiveness of the constructed skills, we do not install \LEGO{} skills into the Codex directory, which allows us to test whether the model can autonomously use the constructed skills.
All experiments use Codex CLI version 0.98.0 as the code agent platform with gpt-5.2-codex and extra-high (xhigh) reasoning effort. Every interaction with the LLM is performed directly through Codex CLI rather than API calls.

Our benchmark uses the specification-to-RTL tasks from VerilogEval v2~\cite{pinckney2024revisitingverilogevalnewerllms}.
Unlike VerilogEval v1~\cite{liu_verilogeval_2023}, VerilogEval v2 includes only HumanEval style problems and emphasizes one-shot success rate Pass@1. This better aligns with the single generation setting in practical engineering.
To better match practical engineering scenarios and more clearly reveal the performance gains achieved by \LEGO{} skills, we filter the dataset.
VerilogEval v2 contains 156 specification-to-RTL tasks, and a large portion of them can be solved directly by modern LLMs.
We run end-to-end generation on VerilogEval v2 with gpt-5.2-codex and select the 41 difficult tasks that gpt-5.2-codex fails to solve as the benchmark for this work.
Following prior work~\cite{11133191,Ho_Ren_Khailany_2025,11105957,11294897}, we compute Pass@1 as
\begin{equation}
\mathrm{pass@}k = \mathbb{E}_{\mathrm{Problems}} \left[1 - \frac{\binom{n-c_p}{k}}{\binom{n}{k}}\right]
\end{equation}
where $k=1$ and $c_p$ is the number of passing runs.
This Pass@1 metric accounts for multiple runs and reflects the expected percentage of problems that the system solves correctly when executed once for each problem.
To evaluate skill composition flexibility, Experiment 2 constructs cross-project combinations that cover diverse design methodologies such as hierarchical methods, specification-driven methods, and understanding-based methods. It also covers complete workflow stages such as specification generation, RTL generation, and debugging. This ensures fair comparison with baseline works.

Tables~\ref{tab:exp1-summary} and \ref{tab:exp2-summary} report solved-problem counts and Pass@1 for Experiments 1 and 2, respectively.
Figure~\ref{fig:exp1-heatmap} visualizes correctness for each problem across the same settings in a grouped layout, with problem IDs on the horizontal axis explicitly showing every problem ID because the benchmark IDs are not consecutive.
For settings with iterative execution, the scheme label includes a second line ``with loop'', and overlaid numbers indicate nonzero loop counts for the corresponding problems.

In Experiment 2, the row and column corresponding to the prior work comparison method hierarchy-verilog are highlighted with a dedicated color to separate external comparison targets from compositions built in \LEGO{}.

\subsection{Workflow Decomposition and Skill Construction}\label{subsec:exp-step-skill}

To verify the effectiveness of the step decomposition and circuit skill construction, Experiment 1 adds one step-level skill group at a time and compares each setting with the baseline gpt-5.2-codex under xhigh. Because the 41 task benchmark is selected from problems that the baseline fails in a previous end-to-end run, the baseline accuracy is zero. As reported in Table~\ref{tab:exp1-summary}, adding G1 and G2 as RTL Generation skills improves Pass@1 by 56.1\% and 61.0\% over the baseline.

We then add S1 from the RTL Spec step on top of the RTL Generation setting to form S1G1. This setting solves one fewer problem than G1. A likely reason is that most tasks in VerilogEval v2 are single module problems with limited need for complex specification decomposition, while the RTL Generation skills already perform lightweight prompt adaptation. Finally, we add the loop-enabled settings E1E2 and E3. They yield clear gains of 70.7\% and 80.5\% over the baseline. The performance improvement from E2 and E3 is largely attributed to Agent Skill RAG, which provides structured debugging experience during the ReAct loop. Figure~\ref{fig:exp1-heatmap} shows that the overlaid loop counts are generally small. This indicates that most problems need only a few iterations to reach correct solutions. The solved problems are distributed across the entire problem ID range instead of concentrating in the lower IDs. This result shows that the constructed circuit skills, the decoupled workflow steps, and the ReAct loop mechanism are effective for solving problems of varying difficulty.

\subsection{Cross Project Skill Composition Comparative Evaluation}\label{subsec:exp-flexibility}

\begin{figure*}[t]
\centering
\includegraphics[width=0.85\textwidth]{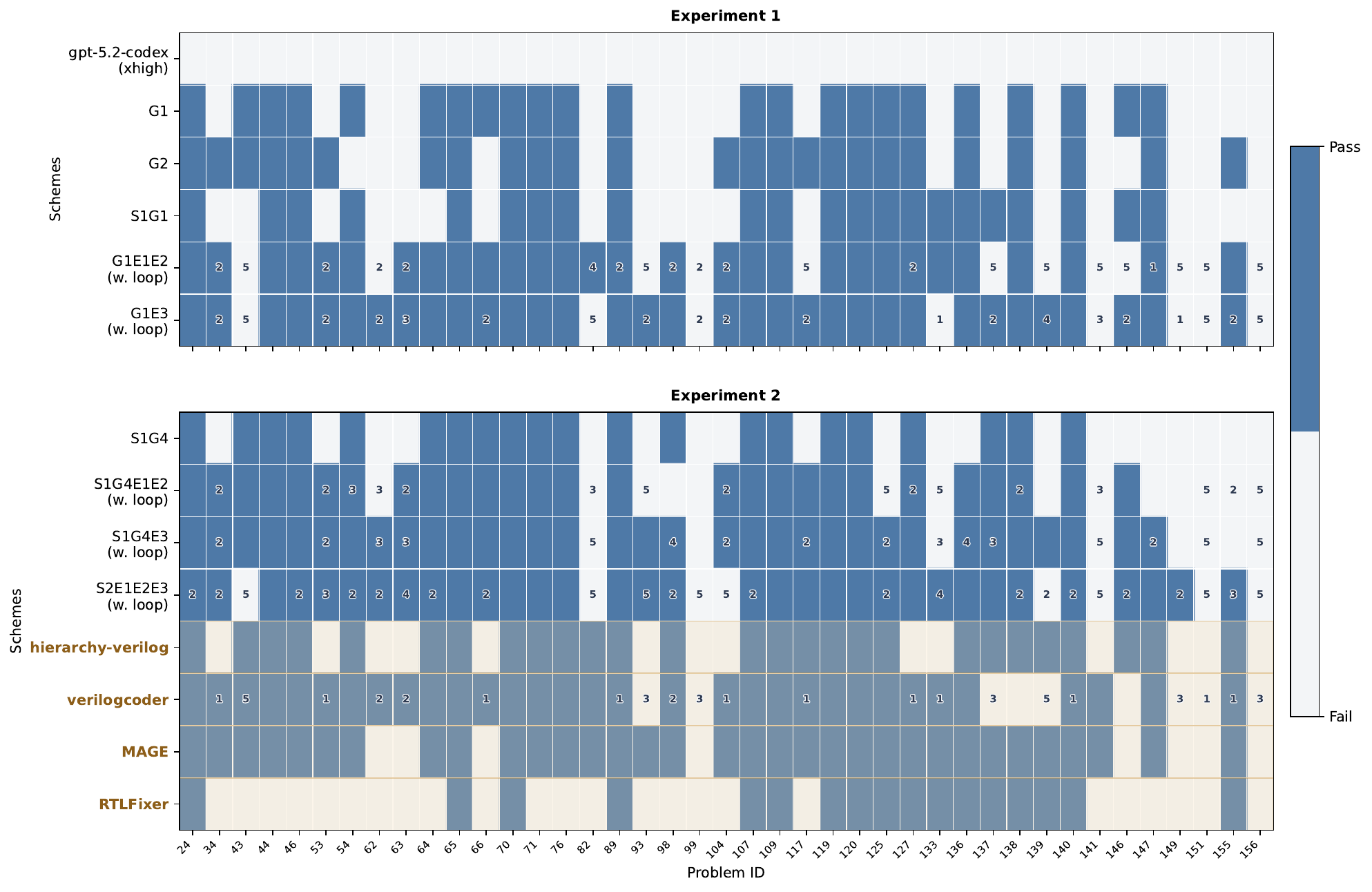}
\caption{Per problem results of Experiments 1 and 2 on 41 filtered VerilogEval v2 problems. The heatmap is split into two aligned subplots for Experiment 1 and Experiment 2. Binary correctness is shown for all settings with dark blue for pass and light gray for fail. All problem IDs are explicitly shown on the horizontal axis. Settings with iterative execution are marked with ``with loop'' on the second line of the scheme label, and overlaid numbers indicate nonzero loop counts.}
\label{fig:exp1-heatmap}
\end{figure*}

To validate the flexibility and effectiveness of modular skill composition, we construct representative combinations that correspond to recognizable design methodologies from baseline works.
S1G4 combines the specification decomposition of hierarchy-verilog, denoted as S1, with hierarchical generation, denoted as G4. It represents the complete hierarchy-verilog workflow~\cite{11105878}.
S1G4E1E2 and S1G4E3 augment this baseline with debugging capabilities from spec2rtl~\cite{11105957} and RTLFixer~\cite{11294897}, testing whether advanced iterative repair can enhance hierarchical methods.
S2E1E2E3 starts from the understanding pipeline of spec2rtl, denoted as S2, and adds comprehensive debugging through E1E2E3. This setting represents an understanding first methodology.
These combinations ensure functional complementarity across workflow stages while enabling direct comparison with their source projects.

We use S1G4 as the internal baseline for measuring composition gains and include hierarchy-verilog~\cite{11105878}, VerilogCoder~\cite{Ho_Ren_Khailany_2025}, and MAGE~\cite{11133191} as external prior work comparisons. The S1G4 baseline already improves Pass@1 by 51.2\% over gpt-5.2-codex under xhigh. We then evaluate multiple cross-project skill compositions. As shown in Table~\ref{tab:exp2-summary}, S1G4E1E2 and S1G4E3 add loop-enabled combinations from spec2rtl~\cite{11105957} and RTLFixer~\cite{11294897} on top of S1G4. They improve by 22.0\% and 29.3\% over S1G4. Another high performing cross source combination is S2E1E2E3. It combines the understanding pipeline of spec2rtl with comprehensive debugging capabilities and improves by 29.3\% over S1G4.

Comparing the composed skills with the original works, S1G4E1E2 outperforms hierarchy-verilog~\cite{11105878} by 7.3\%. S1G4E3 and S2E1E2E3 both outperform hierarchy-verilog by 14.6\% and VerilogCoder~\cite{Ho_Ren_Khailany_2025} by 2.5\%. They also match MAGE~\cite{11133191}. In addition, they exceed standalone RTLFixer~\cite{11294897} by 39.0\%. In isolation, RTLFixer is constrained by a single tool local repair loop and incomplete task context. It often produces syntactic patches that violate behavioral constraints. When integrated into \LEGO{}, upstream decomposition and retrieval supply targeted guidance, while cross skill constraint checking prunes invalid fixes. This turns open-ended repair into guided correction. The multiskill feedback loop includes failure attribution, retry scheduling, and patch validation. It improves stability and explains the large empirical gain.

Figure~\ref{fig:exp1-heatmap} shows that each composed setting corrects a different subset of previously failed problems, further confirming the complementary nature of skills from different sources.

\section{Conclusion}
This paper introduces \LEGO{}, a unified skill-based platform for digital front-end design generation. It decomposes the workflow into six-steps and implements a three-layer architecture that combines orchestration, step skills, and atomic circuit skills. This design enables reusable, configurable, and extensible integration of capabilities from heterogeneous EDA agent projects. Empirical results on 41 hard VerilogEval v2 problems demonstrate both effectiveness and flexibility. In Experiment 1, single skill settings consistently improve over the baseline. Loop enabled settings raise Pass@1 from 0.000 to 0.805 and deliver up to 80.5\% gain over gpt-5.2-codex under xhigh. In Experiment 2, cross-project skill compositions also reach 0.805 Pass@1. They provide up to 29.3\% gain over the S1G4 baseline, outperform hierarchy-verilog by 14.6\%, outperform VerilogCoder by 2.5\%, and match MAGE. These findings show that modular skill construction is a practical direction for robust RTL automation. They also show that circuit skills built in \LEGO{} can closely match and sometimes surpass the performance of original works through flexible composition.

\section{Future Work}

As LLMs advance with stronger agentic capabilities, major platforms including Codex CLI~\cite{openai_codex_cli}, Claude Code~\cite{anthropic_claude_code}, and OpenCode~\cite{opencode} have adopted skill-based architectures. Marketplaces like Skills.homes~\cite{skills_homes} now host thousands of daily updated skills.
The skill paradigm offers natural advantages for EDA through standardization for cross-project reuse, composability for flexible workflows, and text-based interfaces that align with document heavy chip design tasks.


\section*{Acknowledgment}
This project is supported in part by National Science and Technology Major Project (2021ZD0114702).

\end{document}